\documentclass[conference]{IEEEtran}
\IEEEoverridecommandlockouts
\usepackage{cite}
\usepackage{amsmath,amssymb,amsfonts}
\usepackage{algorithmic}
\usepackage{graphicx}
\usepackage{textcomp}
\usepackage{xcolor,colortbl}
\def\BibTeX{{\rm B\kern-.05em{\sc i\kern-.025em b}\kern-.08em
    T\kern-.1667em\lower.7ex\hbox{E}\kern-.125emX}}
\begin{document}

\title{Examining the Effect of Implementation Factors on Deep Learning Reproducibility\\}

\author{
    \IEEEauthorblockN{
        Kevin Coakley\IEEEauthorrefmark{1}\IEEEauthorrefmark{2}, 
        Christine R. Kirkpatrick\IEEEauthorrefmark{1},
        Odd Erik Gundersen\IEEEauthorrefmark{2}
        }
    \IEEEauthorblockA{
        \IEEEauthorrefmark{1}San Diego Supercomputer Center, Univ. of California San Diego, La Jolla, USA
        }
    \IEEEauthorblockA{
        \IEEEauthorrefmark{2}Norwegian University of Science and Technology, Trondheim, Norway
        }
    Email: kcoakley@sdsc.edu, christine@sdsc.edu, odderik@ntnu.no
}

\maketitle

\begin{abstract}
Reproducing published deep learning papers to validate their conclusions can be difficult due to sources of irreproducibility. We investigate the impact that implementation factors have on the results and how they affect reproducibility of deep learning studies. Three deep learning experiments were ran five times each on 13 different hardware environments and four different software environments. The analysis of the 780 combined results showed that there was a greater than 6\% accuracy range on the same deterministic examples introduced from hardware or software environment variations alone. To account for these implementation factors, researchers should run their experiments multiple times in different hardware and software environments to verify their conclusions are not affected.
\end{abstract}


\section{Introduction}

Artificial Intelligence has a reproducibility crisis, much like psychology, medicine, and other fields over the past decade\cite{hutson_2018}. A study of 255 machine learning (ML) papers from 2012 to 2017 showed that only 63.5\% were reproducible\cite{raff_2019}. There are many factors for why papers cannot be reproduced. Gundersen et al. divides the sources that cause irreproducibility into six categories: study design factors, algorithmic factors, implementation factors, observation factors, evaluation factors, and documentation factors\cite{gundersen_2022}. To help with the reproducibility crisis, the effect of these sources of irreproducibility on the conclusions of ML research needs to be studied. We focus on the implementation factors of deep learning (DL) experiments.

\section{Reproducibility}

Reproducibility, a central concept of the scientific method, is often poorly understood. Terms like repeatability and reproducibility are both used interchangeably, even though they have different definitions and have different implications for the conclusions of an experiment. 

Using the definitions from analytical chemistry\cite{miller_2018}:

\begin{itemize}
\item \textbf{Repeatability} is when an experiment is carried out by the same staff, in the same lab, with the same equipment, using the same procedure, within a short period of time.
\item \textbf{Reproducibility} is when an experiment is carried out by different staff, in a different lab, with different equipment, using the same procedure (or as close as possible), at a different time.
\end{itemize}

The lab in analytical chemistry is equivalent to the hardware and software environment in computing. When a deterministic experiment is repeated using the same hardware and software, the results should be the same and therefore the experiment is outcomes reproducible. However, when an experiment is reproduced using a different hardware and software environment by different researchers the outcomes can be different. If the same analysis of the outcomes leads to the same conclusion then the experiment is analysis reproducible. If neither the outcomes nor the analysis are the same as the original, but the interpretation of the analysis leads to the same conclusion then the experiment is interpretation reproducible\cite{gundersen_2021}.

\section{Implementation Factors}

The hardware and software environments used in ML experiments are implementation factors that can contribute to irreproducible results. The following implementation factors are sources of irreproducibility: ancillary software, ancillary software versions, auto-selection of primitive operations, software bugs, compiler settings, initialization seeds, non-deterministic ordering of floating-point operations, parallel execution, and processing unit\cite{gundersen_2022}.

Not all of the implementation factors can be easily controlled for. However, the effects of the implementation factors: ancillary software, ancillary software versions, software bugs, compiler settings, and processing unit \textemdash can be examined by executing multiple runs of the experiment in multiple hardware and software environments. 

\section{Examining the Effects of Hardware and Software Environments on Deep Learning Results}

\subsection{Experiment Setup}

\begin{table*}[t]
\caption{Container Major Software Versions}
\begin{center}
\begin{tabular}{|c|c|c|c|c|c|}
\hline
Container Name & Operating System & Python & TensorFlow & CUDA & cuDNN \\
\hline
tensorflow/tensorflow:2.8.0 & Ubuntu 20.04.3 & 3.8.10 & 2.8.0 & 11.2 & 8.1.0 \\
\hline
tensorflow/tensorflow:2.9.1 & Ubuntu 20.04.4 & 3.8.10 & 2.9.1 & 11.2 & 8.1.0 \\
\hline
nvcr.io/nvidia/tensorflow:22:03 & Ubuntu 20.04.4 & 3.8.10 & 2.8.0 & 11.6.1 & 8.3.3 \\
\hline
nvcr.io/nvidia/tensorflow:22:06 & Ubuntu 20.04.4 & 3.8.10 & 2.9.1 & 11.7 & 8.4.1 \\
\hline
\end{tabular}
\label{tab}
\end{center}
\end{table*}

The following three DL examples were chosen from the Keras GitHub repository\footnote{https://github.com/keras-team/keras-io/blob/master/examples/}: 1) Simple convolutional neural network classifying handwritten digits using MNIST (simple MNIST), 2) a 2-layer bidirectional LSTM classifying sentiments utilizing the IMDB movie review dataset (bidirectional LSTM), 3) a dense neural network performing binary classification on the Kaggle Credit Card Fraud Detection dataset (binary classification). These examples represent three different disciplines within DL, computer vision, natural language processing, and structured data, respectively. They use the TensorFlow library and were modified to run deterministically. The examples were ran five times in each hardware and software environment to verify the deterministic results.

The hardware environments used consisted of resources from the Open Science Grid (OSG). The OSG was chosen because it has many heterogeneous hardware environments. Based on the resources available, runs were completed with four different Intel CPUs, six different AMD CPUs and three different NVIDIA GPUs. The only controlling factor for the placement of the runs was the type of processor available.

Docker containers were used to keep the software environment common across the heterogeneous hardware environments. The four containers that were used are listed in Table I. The tensorflow containers were developed by the TensorFlow team and the nvcr.io containers developed by NVIDIA. The Docker containers were converted to Singularity containers to run on the OSG.

\subsection{Results}

The results showed a greater than 6\% accuracy range of the bidirectional LSTM and binary classification examples when varying the hardware and software environments. The simple MNIST example didn't show significant accuracy differences as all the runs returned a greater than 99\% accuracy. 

The accuracy range of the bidirectional LSTM example exceeded 8\% using the tensorflow/tensorflow containers on GPU hardware; the cause was a software bug (see Conclusions). The bug in the verison of cuDNN included in the tensorflow/tensorflow containers caused runs using a Bi-directional RNN network to return non-deterministic outputs\footnote{https://docs.nvidia.com/deeplearning/cudnn/release-notes/rel\_8.html\#rel-821}. Rebuilding the tensorflow/tensorflow:2.9.1-gpu container with CUDA 11.3 and cuDNN 8.2.1 and rerunning the bidirectional LSTM example eliminated the software bug and produced deterministic results for all three examples.  

\section{Conclusions}

Running the same DL experiment in different hardware and software environments can produce results that are varied enough that a researcher may come to a different conclusion than they would have if they had only ran the experiment in one environment. Hence, claims about the relative performance of DL algorithms such as "DL model A is better than DL model B on task A" cannot be fully trusted if the experiments have not been executed on several hardware and software environments.

When selecting which Keras examples to test, it wasn't known that the tensorflow/tensorflow containers had a version of the cuDNN with a software bug that would affect the results when ran on a GPU. This was discovered only when we ran the bidirectional LSTM example multiple times in different software environments while controlling for other ML implementation factors. A researcher may not expect a TensorFlow Docker container released by the TensorFlow team in July 2022 to have a software bug that was fixed by NVIDIA in June 2021. 

ML researchers should not focus on the variation in the outcomes but how the variation affects the analysis and interpretation. Borrowing again from analytical chemistry, international bodies recommend reproducing experiments in at least 8 different labs. We suggest a similar strategy for ML experiments that rely on stochasticity and floating point calculations, which is the case for DL algorithms. However, exactly what the strategy should be is the objective of future work. These three examples were outcomes reproducible only when deterministic and using the same hardware and software environment. Researchers should test their hypothesis on multiple hardware and software environments to validate their experiments are analysis or interpretation reproducible. 

\section*{Acknowledgment}

This research was done using services provided by the OSG Consortium \cite{osg_1}\cite{osg_2}, which is supported by the National Science Foundation awards \#2030508 and \#1836650.

\bibliographystyle{IEEEtran}
\bibliography{references}

\end{document}